\documentclass[twocolumn]{svjour3}          
\smartqed  

\usepackage{multirow}
\usepackage{adjustbox}
\usepackage{graphics}
\usepackage{array}
\usepackage{arydshln}
\usepackage{tabularx}
\usepackage{tikz}
\usepackage{pgfplots}
\pgfplotsset{width=0.5\textwidth,compat=1.9} 
\begin{document}
\title{Human-Robot Handshaking: A Review}
%
%
\author{Vignesh Prasad$^1$
\and Ruth Stock-Homburg$^1$
\and Jan Peters$^{1,2}$}%
\authorrunning{V. Prasad et al.}

\institute{ $^1$ Technical University of Darmstadt \\
              Hochschulstr. 1\\
              64289 Darmstadt, Germany\\\\
           $^2$ Max Planck Institute for Intelligent Systems, T\"ubingen, Germany
    \\\\
    Contact Author: Vignesh Prasad\\
    Tel.: +49 6151 16-24475\\
    Fax: +49 6151 16-24460\\
    \email{vignesh.prasad@tu-darmstadt.de}
}
\maketitle              
\begin{abstract}
For some years now, the use of social, anthropomorphic robots in various situations has been on the rise. These are robots developed to interact with humans and are equipped with corresponding extremities. They already support human users in various industries, such as retail, gastronomy, hotels, education and healthcare. During such Human-Robot Interaction (HRI) scenarios, physical touch plays a central role in the various applications of social robots as interactive non-verbal behaviour is a key factor in making the interaction more natural. Shaking hands is a simple, natural interaction used commonly in many social contexts and is seen as a symbol of greeting, farewell and congratulations. 
In this paper, we take a look at the existing state of Human-Robot Handshaking research, categorise the works based on their focus areas, draw out the major findings of these areas while analysing their pitfalls. We mainly see that some form of synchronisation exists during the different phases of the interaction. In addition to this, we also find that additional factors like gaze, voice facial expressions etc. can affect the perception of a robotic handshake and that internal factors like personality and mood can affect the way in which handshaking behaviours are executed by humans. Based on the findings and insights, we finally discuss possible ways forward for research on such physically interactive behaviours.
\end{abstract}
\section{Introduction}
\label{sec:intro}
In the context of Social Robots and Human-Robot Interaction (HRI), there has been an increase in the use of anthropomorphic robots in social settings, and they already support human users in various industries, such as retail, gastronomy, hotels, education, and healthcare services \cite{feil2005defining,han2012robotic,jeffrey2016meet,zhang2008service}. In HRI, physical presence plays an important role as it can influence the image of a robot as compared to a virtual presence \cite{li2015benefit}.  According to the Computer-As-Social-Actor (CASA) paradigm \cite{nass2000machines}, physical presence is also an important predictor of the mindless social response to robots by which humans put a robot in the same category as humans. 
These responses are supposedly triggered by social human-like cues as well, such as voice \cite{nass1997machines},  face \cite{nass2001truth}, and language style \cite{nass1995can}. Additionally, the appearance of the robot, as opposed to just the physical presence, can also have an impact on the perception of robots. As the Uncanny Valley \cite{mori1970uncanny} hypothesises, uncanny feelings are triggered by highly human-like robots (androids) but not as realistic as a human being as compared to robots that are less human-like in appearance (humanoids). These uncanny feelings are hypothesised to get exaggerated even more in the case of movements. The affinity, of both appearance and movement, rises again only when the human-likeness becomes very close to that of a human being.

\begin{figure*}
\centering
    \includegraphics[width=0.8\textwidth]{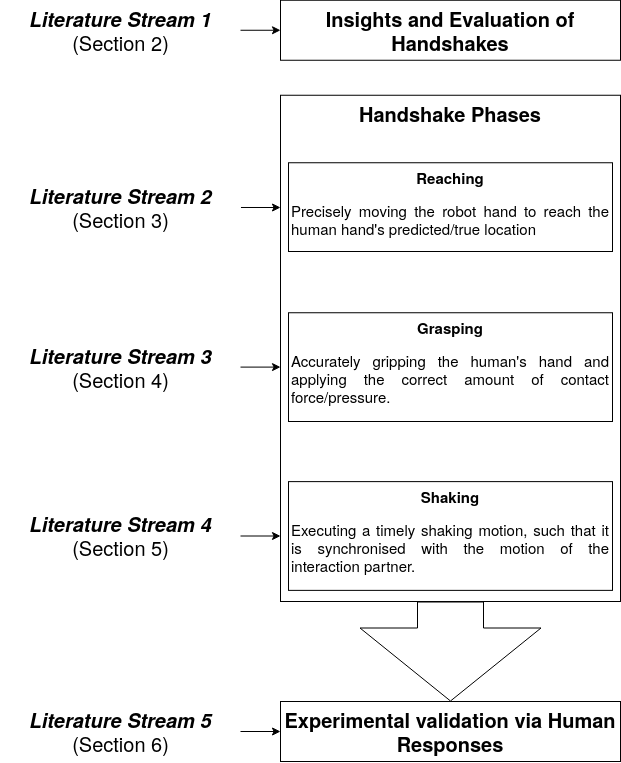}
    \caption{Conceptual Framework for categorising works on Human-Robot Handshaking}
    \label{fig:framework}
\end{figure*}

\begin{figure}
    \centering
    \begin{tikzpicture}  
  
    \begin{axis}  
    [  
        title={\textbf{Publications on Human-Robot Handshaking}},
        ybar,
        bar width=27pt,
        enlargelimits=0.12,  
        ylabel={\textbf{Number of Publication}}, 
        xlabel={\textbf{Years}},  
        symbolic x coords={1996-2000, 2001-2005, 2006-2010, 2011-2015, 2016-2020}, 
        xtick=data,  
         nodes near coords, 
        nodes near coords align={vertical},  
        ]  
    \addplot coordinates {(1996-2000,1) (2001-2005,1) (2006-2010,12) (2011-2015,20) (2016-2020,27)};  
    \end{axis}  
    \end{tikzpicture}    
    \caption{Publications on Human-Robot Handshaking}
    \label{fig:pub_year}
\end{figure}
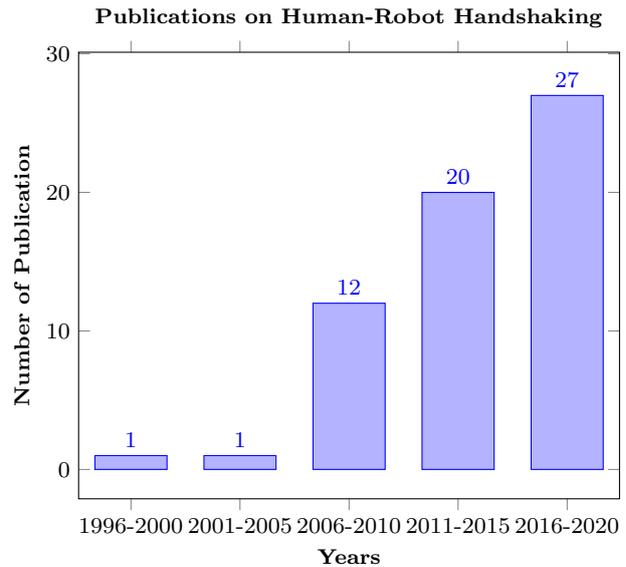

In the context of HRI, physical contact plays a central role in the various applications of social robots. One of the key reasons for this being that non-verbal behaviour, especially touch, can be used to convey information about the emotional state of a person \cite{hertenstein2006communicative,ogden2000robotic,yohanan2012role}. This enables a special kind of emotional connection to human users during the interaction \cite{han2012robotic,stock-homburg2018negative}. If, for example, one considers a future scenario in which an accompanying robot shares the habitat with humans, a key requirement for the robot would be its ability to physically interact with humans \cite{wang2018predicting}. In such a case, it would be advantageous that humans feel more welcoming and be willing to interact and help a social robot with its task like they would help other humans. Among such interactions, handshaking is a common natural physical interaction and an important social behaviour between two people \cite{schiffrin1974handwork}, that is used in different social contexts \cite{chaplin2000handshaking,stewart2008exploring,hall1983handshake}. The importance of touch in HRI and the prolonged nature of the contact additionally make handshaking a more important interaction as compared to other interactions, like high fives or other Asian greeting behaviours which do not involve physical contact.  Handshaking can, therefore, represent an important social cue according to the CASA paradigm for several reasons:

\begin{itemize}
    
    \item It is one of the first and foremost non-verbal interactions which takes place and should, therefore, be part of the repertoire of a social robot.
    
    \item It plays an important role in shaping the impressions of others \cite{chaplin2000handshaking,stewart2008exploring,bernieri2011influence}, which is used to develop an initial personality judgement of a person \cite{aastrom1996greeting,aastrom1994introductory}. 

     \item The shaking of hands is seen as a symbol of greeting, farewell, agreement, or congratulation. Socially, it symbolises acceptance and respect for another person \cite{papageorgiou2015kinematic}. The most common of these settings is "greeting" in which it is usually the first non-verbal interaction taking place in a social context.

    \item The shaking of hands may help set the tone of any interaction, especially since the sense of touch can convey distinct emotions \cite{hertenstein2006communicative}.  

    \item A good robot handshake may lead to future cooperation and coexistence \cite{papageorgiou2015kinematic}. 
\end{itemize}
This can also be further enriched, such as in the possible scenario wherein a robot can monitor the biological attributes of a person (such as stress levels form blood flow) and thereby infer social information about a person from just a single handshake \cite{duffy2003anthropomorphism}. Having human-like body movements plays an important role in the acceptance of HRI wherein humans tend to look at robots more as social interaction partners \cite{kupferberg2011biological}. In the case of humanoid robots, having realistic motions enable similar responses as humans \cite{chaminade2005motor}. Thus, having a good handshake can not only widen the expressive abilities of a social robot but also provide a strong first impression for further interactions to take place. Robot handshaking can additionally help improve the perception of the robot and enable humans to be more willing to help a robot \cite{avelino2018power} allowing for better integration of the robot into human spaces. To perform proper handshaking motions, a social robot should be able to detect and predict movements of the human and react naturally. Therefore, for better acceptance and improved expressiveness, effective handshaking behaviours need to be present to make social robots feel more acceptable. This importance can be seen in Fig. \ref{fig:pub_year}, in the rising trend of works on  human-robot handshaking.

Now that the importance of handshaking from a robotic standpoint has been established, we propose the following framework for categorising the different works, shown in Fig. \ref{fig:framework}. We first discuss works that aim to model handshakes from human-human interactions. Along with that, we discuss the evaluation criteria used for analysing participants' feedback in experiments with humans interacting with the robot. This can be seen in Sec. \ref{sec:modelling}. Following this, we go through each of the stages of handshaking, as depicted in \cite{melnyk2014analysis,vinayavekhin2017human}, namely reaching (Sec. \ref{sec:reaching}), grasping (Sec. \ref{sec:grasping}) and shaking (Sec. \ref{sec:shaking}). We then explore more into the responses of participants to different handshaking methods along with various factors, such as their acceptability, their degree of preference, how external factors like voice and gaze affect the perception of robot handshakes and so on in Sec. \ref{sec:social}. Finally, we discuss some of the shortcomings of existing works and propose some areas for future research in Sec. \ref{sec:discussion} and present our concluding remarks in Sec. \ref{sec:conc}. This was done as a broad categorisation of the works obtained after a search of digital libraries, Google Scholar, IEEE Xplore and ACM Digital Library using keywords that included "human-robot handshaking", ”robotic handshaking” and ”handshaking AND human-robot interaction / HRI” followed by a depth-first search-styled approach among the references and citations of the papers found. Of these, we included all the articles that were published in conferences or journals. We build upon our previous work \cite{prasad2020advances} where we talked about the different works that model human handshaking interactions, and the works in the different stages of the handshaking exchange and some social responses of participants. In this paper, we dig deeper into these aspects and provide key findings and insights of the different areas of work in regards to Human-Robot Handshaking. In the end, we propose some ways forward based on aspects that still need to be worked on in order to realise a social robot that can truly capture the intricacies of such a physically interactive action.

\section{Insights and Evaluation of Handshakes}
\label{sec:modelling}
Before going into robotic handshakes, we first discuss a few works that draw insights from human handshakes and some evaluation mechanisms used to measure the parameters related to human acceptance of robotic handshakes. Regarding the insights drawn, the main parameters looked at include trajectory profiles (mainly acceleration, velocity, contact forces) between participants when they shake hands and the mutual synchronisation of their movements while doing so. Regarding the evaluation criteria, they mainly relate to how human-like the handshake is and how the handshakes are perceived by humans who interact with the robot.

\subsection{Insights from Human-Human Handshake Interactions}
A group of researchers from Okayama Prefectural University, Japan conducted a series of studies \cite{jindai2015handshake,jindai2012small,jindai2007development,jindai2008handshake,jindai2010small,jindai2011development,jindai2006development,yamato2008development,ota2014handshake,ota2015handshake} to study handshaking interactions between humans and analyse how participants respond to different behaviours applied on a custom robotic arm. They use a VICON motion capture system to track markers placed on both shoulders, right elbow wrists and hands of both participants. 
Initially, Jindai et al. \cite{jindai2006development} observed that the motion of the responder was seen to be similar to that of the requester with a lag between their velocity profiles, which was found to be similar to a minimum jerk trajectory profile. Hence they applied a lag-based transfer function for generating the robot's reaching trajectory based on the human's reaching motion. It was previously shown that this type of "motion transfer" was emotionally acceptable to humans in an object-handover scenario \cite{yamamoto2005application}. Subsequently, the oscillatory motion profiles of the observed shaking behaviours were modelled as a spring and damper system \cite{jindai2008handshake,yamato2008development}. Following this, the interactiveness was studied by modelling the requester\cite{jindai2011development,jindai2015handshake} and responder\cite{ota2014handshake,ota2015handshake} behaviours with respect to delays in the motions and auxiliary behaviours like gaze and speech. The result of applying the key findings from the above works on their robot is explored in Sec. \ref{sec:social}.
Their main findings of the modelling can be summarised as follows.
\begin{itemize}
    \item The reaching trajectories of the requestor and responder are similar in their trajectory profiles, which follow a minimum jerk trajectory model in \cite{jindai2011development}, and the motion of one can be used to mimic the other, such as by using a transfer function \cite{jindai2006development}. 
    \item As expected, the shaking behaviour and the transition from reaching to shaking is modelled as a spring-damper system, given the oscillatory nature of the behaviour \cite{yamato2008development}. 
    \item Leading handshake behaviours were preferred over a non-leading one \cite{yamato2008development} and a small delay (0.2s - 0.4s) between responding to a handshake request was better perceived. 
    \item In terms of auxiliary behaviours, using voice with a small or no delay was preferred and having the gaze shift steadily from the hand to the face was well perceived.
\end{itemize}

A group of researchers from the University of Lorraine, France explicitly study the mutual synchronisation (MS in short) between participants while shaking hands along with the forces exerted on the palms \cite{melnyk2014analysis,melnyk2019physical,melnyk2014sensor,melnyk2015analysis,tagne2016measurement}. They firstly study the hand motions by having the participants wear a glove with an Intertial Measurement Unit (IMU) and 6 force sensors placed around the palm \cite{melnyk2014sensor}. Tagne et al. \cite{tagne2016measurement} further investigate the joint motions as well (elbow and shoulder) with IMUs placed at each joint. The influences of a few different social settings, such as greeting, congratulating or sympathy, are then explored as well \cite{melnyk2019physical,tagne2016measurement}. The MS between participants is analysed using the Fourier analysis of the input signals (mainly accelerations). However, wavelet transforms are shown to qualitatively estimate the different stages of a handshake interaction as well \cite{melnyk2015analysis}. 

Initially, the MS between participants was explored in a context-less setting along with the contact strength of the interaction \cite{melnyk2014analysis}. The mean duration of a handshake was around 2.67 $\pm$ 0.87 seconds during which the average duration of grasping was similar across pairs (~ 0.5 s) whereas the duration of shaking had a larger variation, from just below 1 s to almost 2.5 s. The frequency of shaking during MS peaked around 4Hz. The average strength of the contact, which is the average of the forces measured by the sensors was 2.5N (no standard deviation was reported). This framework was extended to analyse differences in a few social contexts \cite{melnyk2019physical,tagne2016measurement} and gender-based differences\cite{melnyk2019physical}. Tagne et al. \cite{tagne2016measurement} observe the differences between 3 scenarios namely greeting, sympathy and congratulating. Melnyk and H\'enaff \cite{melnyk2019physical} analyse similar social settings, namely greeting and consolation, and additionally analyse the trends across different gender-based pairings.

\begin{table*}[]
    \centering
\resizebox{\textwidth}{!}{%
\begin{tabular}{|c|c|c|c|c|c|}
\hline
\multirow{2}{*}{Study}                  & \multicolumn{2}{c|}{\multirow{2}{*}{Setting}} & Duration    & Wrist Frequency              & Grip Strength \\
                                        & \multicolumn{2}{c|}{}                         & (s)         & during MS (Hz)                              & (N)           \\ \hline
Melnyk et al.  \cite{melnyk2014analysis}                    & \multicolumn{2}{c|}{None}           & $2.67 \pm 0.86$ & $4.20 \pm 1.00$                       & $2.50$          \\ \hline
\multirow{3}{*}{Tagne et al. \cite{tagne2016measurement}}      & \multicolumn{2}{c|}{Greeting}       & $0.90 \pm 0.26$ & $2.43 \pm 0.62$                       & $4.37 \pm 2.4$    \\ \cdashline{2-6} 
                                        & \multicolumn{2}{c|}{Sympathy}       & $1.30 \pm 0.49$ & $2.44\pm0.69$                         & $3.12 \pm 2.1$    \\ \cdashline{2-6} 
                                        & \multicolumn{2}{c|}{Congratulation} & $1.24 \pm 0.40$ & $2.66\pm0.72$                         & $5.88 \pm 3.2$    \\ \hline
\multirow{6}{*}{Melnyk and H\'enaff \cite{melnyk2019physical}} & \multirow{3}{*}{Greeting}     & MM  & $0.73 \pm 0.08$ & \multirow{3}{*}{3.6 (median 3.5)} & $6.02 \pm 0.99$   \\ \cdashline{3-4} \cdashline{6-6} 
                                        &                               & MF  & $1.48 \pm 0.40$ &                                   & $5.68 \pm 0.8$    \\ \cdashline{3-4} \cdashline{6-6} 
                                        &                               & FF  & $1.95 \pm 0.26$ &                                   & $5.53 \pm 1.17$   \\ \cdashline{2-6} 
                                        & \multirow{3}{*}{Consolation}  & MM  & $2.40 \pm 0.23$ & \multirow{3}{*}{3.7 (median 3.5)}  & $6.08 \pm 1.04$   \\ \cdashline{3-4} \cdashline{6-6} 
                                        &                               & MF  & $2.54 \pm 0.42$ &                                   & $6.18 \pm 0.97$   \\ \cdashline{3-4} \cdashline{6-6} 
                                        &                               & FF  & $4.05 \pm 0.53$ &                                   & $6.29 \pm 0.85$   \\ \hline
\end{tabular}
}
\caption{Summary of Results of \cite{melnyk2014analysis,melnyk2019physical,tagne2016measurement} (M - Male, F - Female. Values are reported as mean $\pm$ standard deviation and only mean is reported when standard deviation is not available.)}
\label{tab:melnyk}
\end{table*}

As seen in Table \ref{tab:melnyk}, a shorter duration was observed in greeting contexts. The duration in cases of sympathy and congratulations were similar. The grip strength shows contradictory results. Tagne et al. \cite{tagne2016measurement} saw the lowest grip strength in case of sympathy, followed by greeting and then, congratulations. Melnyk and H\'enaff \cite{melnyk2019physical} found that it was slightly higher in consolation case although not significantly. In terms of gender-based pairings, it is seen that MM pairs shook for a lesser duration as compared to mixed pairings. Female pairs shook hands the longest. This is consistent with another study \cite{orefice2016let} as well. No conclusive correlations were found between gender and grip strength, contrary to previous studies\cite{chaplin2000handshaking,orefice2016let}. 

Unlike the above works that explicitly measure the stiffness and forces of the interactions, Dai et al. \cite{dai2019research} indirectly model the stiffness of the elbow joint as a spring-damper system, like few other works described in Sec. \ref{sec:shaking}. They measure the expansion/contractions of the muscles in the upper arm and forearm using EMG signals and thereby estimate the stiffness of the elbow using the biceps and triceps and use the forearm muscle measurements to observe indications of the grasping forces. Data was collected from 10 handshakes of which 5 were weak and 5 were strong. It was very evidently seen that muscle activation in the case of the strong handshake was higher than the weak condition.

The above works mainly study human-human handshaking to gain insights into the motion and forces involved in handshaking. However, works looking into how well robotic interfaces are suited for such haptic-heavy interactions are limited in number. In this regard, Knoop et al. \cite{knoop2017handshakiness} perform experiments to understand the contact area, contact pressure and grasping forces exerted by participants during handshaking and test out how a few robotic hands and custom finger designs comply with their observations from human-human handshaking interactions. Participants were asked to perform 3 handshakes of different strengths, namely, weak, normal and strong. A large variation was seen in the final contact positions of the fingers at the back of the hand. In the front, there is little variation across different handshakes as almost the whole palm is held during the contact. This would imply that there is possibly no fixed grasping location for a handshake, and that the palms should be sufficiently covered. during the interaction. A positive correlation was found between contact pressure and grasping force, which is a straightforward implication since the contact area doesn't vary during the interaction. They test out how a few robotic hands and custom finger designs compare with a human hand and argue that this study is useful for optimising robotic hand designs at a coarse level.

\textit{\textbf{Major Findings}}. Handshaking is inherently a synchronous process, which is observed in the reaching motions by \cite{jindai2006development,jindai2007development} and the shaking by \cite{melnyk2014analysis,melnyk2019physical,tagne2016measurement}. This would imply that both parties involved in a handshake try to achieve a common motion during the action. This kind of inherent similarity in the motions and the subsequent synchronisation can therefore be treated as an important aspect of making a handshaking behaviour more acceptable. The context of a handshake along with additional factors, like speech and gaze, play a role in the interaction as well. The factors studied and the measurements obtained (in terms of duration, frequency, relative grip strength etc.) can further be used to explicitly model robotic handshake behaviours to give them a "personality" of sorts or provide appropriate responses based on detected interaction contexts improving the social understanding of the robot.

\begin{table*}[]
\centering
\begin{tabularx}{\textwidth}{|c|>{\centering\arraybackslash}X|>{\centering\arraybackslash}X|}
\hline
\textbf{Evaluation Method} & \textbf{Works} & \textbf{Evaluation Parameters} \\ \hline
\multirow{2}{*}{Bradley Terry Model} & Jindai et al.\cite{jindai2015handshake,jindai2012small,jindai2007development,jindai2008handshake,jindai2010small,jindai2011development,jindai2006development}, Ota et al. \cite{ota2014handshake,ota2015handshake}, Yamato et al. \cite{yamato2008development} & Participant's preferences of handshakes \\ \cline{2-3} 
 & Kasuga and Hashimoto \cite{kasuga2005human} & Flexibility, naturalness, kindness, affinity \\ \hline
\multirow{3}{*}{Seven point scale} & Jindai et al.\cite{jindai2015handshake,jindai2012small,jindai2007development,jindai2008handshake,jindai2010small,jindai2011development,jindai2006development}, Ota et al. \cite{ota2014handshake,ota2015handshake}, Yamato et al. \cite{yamato2008development} & Handshake motion, Security, Velocity/comfort, politeness/Vitality \\ \cline{2-3} 
 & Avelino et al. \cite{avelino2018power} & RoSaS (Warmth, competence, discomfort), Godspeed (anthropomorphism, animacy, likeability), closeness, willingness to help robot \\ \cline{2-3} 
 & Mura et al. \cite{mura2020role}, Vigni et al. \cite{vigni2019role} & Quality, human-likeness, responsiveness, perceived leader, personality \\ \hline
\multirow{3}{*}{Five point scale} & Ammi et al. \cite{ammi2015haptic}, Tsamalal et al. \cite{tsalamlal2015affective} & Valence, Arousal, Dominance of Visual, haptic and visuohaptic interactions \\ \cline{2-3} 
 & Arns et al. \cite{arns2017design} & Compliance, force feedback, overall haptics \\ \cline{2-3} 
 & Christen et al. \cite{christen2019guided} & Naturalness of video of different simulated interactions \\ \hline
\multirow{2}{*}{Score (out of 10)} & Wang et al. \cite{wang2008modelling,wang2009hmm}, Giannopoulos et al. \cite{giannopoulos2011comparison} & Human likeness rating of Robot handshakes \\ \cline{2-3} 
 & Dai et al. \cite{dai2019research} & Naturalness \\ \hline
Model Human Likeness Grade & Avraham et al. \cite{avraham2012toward}, Karniel et al. \cite{karniel2010turing}, Nisky et al.\cite{nisky2012three} & Human likeness of proposed handshake models \\ \hline
\end{tabularx}
\caption{Methods and Parameters used by different works to evaluate robotic handshaking.}
\label{tab:evaluation}
\end{table*}

\begin{table*}[]
\centering
\begin{tabularx}{\textwidth}{|>{\centering\arraybackslash}m{2.5cm}|>{\centering\arraybackslash}m{2cm}|>{\centering\arraybackslash}m{5.57cm}|>{\centering\arraybackslash}m{5.56cm}|}
\cline{3-4}
\multicolumn{2}{c|}{} & \textbf{With force feedback} & \textbf{Without force feedback} \\ \hline
\multirow{3}{2.5cm}{\textbf{Human-like hand (4/5 finger model)}} & \textbf{Inactive} & Dai et al.\cite{dai2019variable} & Campbell et al. \cite{campbell2019learning}, Jindai et al.\cite{jindai2015handshake,jindai2012small,jindai2007development,jindai2008handshake,jindai2010small,jindai2011development,jindai2006development}, Kasuga and Hasimoto \cite{kasuga2005human}, Knoop et al. \cite{knoop2017handshakiness}, Melnyk and Henaff \cite{melnyk2016bio}, Nakanishi et al. \cite{nakanishi2014remote}, Orefice et al. \cite{orefice2018pressure}, Ota et al. \cite{ota2014handshake,ota2015handshake}, Stock-Homburg et al. \cite{stock2020evaluation}, Vanello \cite{vanello2010neural}, Vinayavekhin et al. \cite{vinayavekhin2017human},  Yamato et al. \cite{yamato2008development}    \\ \cline{2-4} 
& \textbf{Passively controlled} & Arns et al. \cite{arns2017design}, Beaudoin et al. \cite{beaudoin2019haptic}, Ouchi and Hashimoto \cite{ouchi1997handshake}, Pedemonte et al. \cite{pedemonte2016design,pedemonte2017haptic} & \\ \cline{2-4} 
& \textbf{Actively controlled}  & Avelino et al. \cite{avelino2018power,avelino2018towards}, Ammi et al. \cite{ammi2015haptic}, Christen et al.\cite{christen2019guided}*,  Mura et al.\cite{mura2020role}, Tsamalal et al.\cite{tsalamlal2015affective}, Vigni et al.\cite{vigni2019role}  & \\ \hline
\multicolumn{2}{|>{\centering\arraybackslash}c|}{\textbf{Gripper}} & & Bevan and Fraser \cite{bevan2015shaking}, Falahi et al. \cite{falahi2014adaptive}, Jouaiti et al.\cite{jouaiti2018hebbian}*, Sato et al.\cite{sato2007synchronization}* \\ \hline
\multicolumn{2}{|>{\centering\arraybackslash}c|}{\textbf{Rod-like end-effector}} & & Avraham et al.\cite{avraham2012toward}, Giannopoulos et al.\cite{giannopoulos2011comparison},  Karniel et al.\cite{karniel2010turing}, Nisky et al.\cite{nisky2012three}, Papageorgiou and Doulgeri\cite{papageorgiou2015kinematic}, Wang et al.\cite{wang2009hmm,wang2011handshake}  \\ \hline
\end{tabularx}
\caption{Different types of Robot End Effectors used for Human-Robot Handshaking. *-simulated robot}
\label{tab:endeff}
\end{table*}

\subsection{Handshake Evaluation Methods}
Given the differences in hardware and the evaluation criteria used by different works, it is difficult to converge on a single metric or scale for the task at hand. Moreover, different works evaluate different aspects of handshaking, using different robots. It is therefore difficult to come up with a common comparison baseline, although some studies evaluate their methods similarly. To this end, we collate some of the common evaluation metrics and methods used among different works in Table \ref{tab:evaluation} and broadly categorise some of the different robotic interfaces in Table \ref{tab:endeff}. Some common aspects are the aim to rate the acceptability of the handshaking interactions and the human-likeness or the naturalness of the handshaking. Moreover, most of the works that use a human-like end effector mainly have an inactive one, which could cause the interaction to seem more unnatural. For example, in the works of Wang et al. \cite{wang2011handshake} and Giannopoulos et al.\cite{giannopoulos2011comparison} who use a rod-like end-effector, the case when the robot arm is operated by a human, the handshake gets an average human-likeness rating of only 6.8/10 which is far from the maximum score. 

The common metrics used, like in most psychological and human studies, are the seven-point or the five-point Likert scales, which is a bipolar scale that has a negative valued sentiment on one end and a positive valued one on the other, which allows for a nice representation, especially when averaging over the data. For example, an overall negative average indicates an inclination towards the negative sentiment and a positive average indicates an inclination towards the positive sentiment. This, in contrast to comparing an absolute score (rating out of ten for example), can help indicate the sentiments of the participants better. 

To use a more traditional test of computational intelligence, Karniel et al. \cite{karniel2010turing} propose a Turing test for motor intelligence and come up with a metric called as the Modern Human Likeness Grade (MHLG) which is used to indicate the human-likeness of different shaking behaviours in this kind of a mechanical Turing Test. This is based on the perceived probability by a participant of the model being a human shaking the stylus or the algorithm. Nisky et al. \cite{nisky2012three} propose different ways to perform this Turing test for motor intelligence. There are described in further detail in Sec. \ref{ssec:misc_shaking}.

\section{Reaching phase of Handshaking}
\label{sec:reaching}
We have already described the work of Jindai et al. \cite{jindai2015handshake,jindai2011development,jindai2006development} and Ota et al. \cite{ota2014handshake,ota2015handshake} above. To the best of our knowledge, these were the first works to model the hand reaching aspect and deploy it on a robot. As mentioned above they propose two models. One with a transfer function based on the human hand's trajectory with a lag element and the other is a minimum jerk trajectory model, which fits the velocity profiles and provides smooth trajectories by definition. These modelling choices imply that a smooth motion similar to the interaction partner is preferred with a small amount of delay between them. However, they do not have any study showing how these two models compare with each other. 

More recent works model reaching using machine learning. Campbell et al. \cite{campbell2019learning} use imitation learning to learn a joint distribution over the actions of the human and the robot. During testing time, the posterior distribution is inferred from the human's initial motion from which the robot's trajectory is sampled. Their framework estimates the speed of the interaction as well, to match the speed of the human. Christen et al. \cite{christen2019guided} use Deep Reinforcement Learning (RL) to learn physical interactions from human-human interactions. They use an imitation reward which helps in learning the intricacies of the interaction. Falahi et al. \cite{falahi2014adaptive} use one-shot imitation learning to kinesthetically teach reaching and shaking behaviours based on gender and familiarity detected using facial recognition. However, it cannot be generalised due to the extremely low sample size. Vinayavekhin et al. \cite{vinayavekhin2017human} model hand reaching with an LSTM trained using skeleton data. They predict the human hand's final pose and devise a simple controller for the robot arm to reach the predicted location. In terms of smoothness, timeliness and efficiency, their method performs better than following the intermediate hand locations. However, it performs worse than using the true final pose due to inaccuracies in the prediction.

\textbf{\textit{Major Findings}}. Modelling of reaching behaviours draws heavily on learning from human interactions, unlike other robotic grasping/manipulation tasks, where a lot of it can be learnt from scratch. This provides a strong prior to help make the motions more human-like and can also be used to initialise \cite{campbell2019learning} or guide \cite{christen2019guided} the learning. 


\section{Controlling Hand Grasps in Handshaking}
\label{sec:grasping}
One of the first remote handshaking systems, proposed by Ouchi and Hashimoto \cite{ouchi1997handshake}, was aimed at two people performing a handshake while on a telephone call with each other using a custom-made silicone-rubber based robotic soft hand. They measured the pressure exerted on the hand using a pneumatic force sensor which relays the force information over to the other user, who has a robotic hand as well. With this type of active haptic mechanism, they show that users better perceive the partner's existence during the call and that they were able to shake hands without feeling any transmission delay. This shows the effect that such haptic interactions have on the perception of the interaction partner.

Pedemonte et al. \cite{pedemonte2016design} design an anthropomorphic haptic interface for handshaking. It is an under-actuated robot hand with a passive thumb that is controlled based on the amount of force that is applied on it. It is a sensor-less model with a deformable palm that controls the closure of the fingers. A variable admittance controller is used to set the reference position for fingers based on the degree deformation of the palm. Therefore the amount of force exerted by the robot hand on the human hand depends on the force exerted by the human, leading to a partial synchronisation in the grasping. It takes approximately 0.6s to close the fingers. Arns et al. \cite{arns2017design} build upon this design using lower gear ratios and more powerful actuators to obtain a stronger grasping force and a faster interaction speed. They argue that the use of impedance control as opposed to admittance control helps improve responsiveness as well. A similar synchronisation is observed as in the previous work as the mechanisms are the same, in theory. The main difference is the speed of the interaction which is almost instantaneous in this case (less than 0.05s), making the interaction more realtime and natural.

Avelino et al. \cite{avelino2017human,avelino2018towards} propose two models to develop a pleasant grasp for handshaking. This is extended to three different grasping models with different degrees of hand closure, corresponding to a strong, medium strength and a weak handshake \cite{avelino2017human}. Force sensors present on the robots finger joints measure the interaction forces during grasping. It was found that female participants mainly preferred strong handshakes (85.7\%). There was a larger variability among male participants. Since a simple position based control is employed, the force perceived depends on the hand sizes of the participants, which could be the cause of the variability. This is addressed in \cite{avelino2018towards}, where an initial study is carried out where participants have to adjust their hand and the robot's grip until a preferable grasp is reached. This is done to find a suitable reference force distribution among the sensors on the robot hand. The finger joint positions are recorded as well. With this distribution, they compare a fixed handshake to a force control method. The force control is done with a PID controller whose set points are the average of the forces per sensor on each finger obtained from the previous data. Moreover, they combine this with a shaking motion presented in \cite{wang2008modelling} that is described in Sec. \ref{sec:shaking}. Participants had to rate the two handshakes based on various factors like scariness, arousal (boring/interesting), meaningfulness, excitement, strength/firmness, and the perceived enjoyment and safety, all on 7-point scales for each variable. Although both handshakes were evaluated positively overall, no significant differences were observed between them.

Vigni et al. \cite{vigni2019role} model the force exerted by the robot hand during handshaking based on the force exerted by the human, measured using force-sensitive resistors on the robot hand. The robot force is approximated from the degree of hand closure using a calibration experiment where participants are asked to mimic the force felt on their hand by a few open-loop handshakes of the robot. The human force is estimated by fitting a cubic polynomial to the sum of forces applied on the individual sensors. This is also done with a calibration experiment where participants were made to grasp a sensorised palm fitted with a load cell to measure the exerted force. They compare three different controllers based on the relationships between the exerted forces of the human and the robot namely linear, constant and combined (constant+linear). The latter two are used with two values of the constant force, weak and strong. Since humans have a small delay in reaction time, a controller delay of 120ms was observed to be more natural and was added to the behaviour. 

The mean duration of handshakes was 2.2s with 24.8N as the mean sum of forces measured on the robot hand exerted by the humans. The participants (n=15) filled out a survey after interacting, rating the quality, human-likeness, responsiveness, perceived leader, and the perceived personality of the robot on a 7 point scale. The combined controllers were perceived better than the constant ones in terms of quality, human-likeness and responsiveness, with a significant difference between the weak variants. There was no significant effect in terms of who the perceived leader or follower was. However, it was observed that in the constant force cases, humans would adjust their force based on the robot's, showing that humans tend to follow the force exerted on their hand. These findings further emphasise the effect of mutual synchronisation in handshaking. In terms of personality, the stronger variants of the constant and combined controllers were perceived as more confident/extroverted, with a significant effect seen between the variants of the constant controller.

\textbf{\textit{Major Findings}}. The main commonality among the above-mentioned works is that a force feedback mechanism is necessary to ensure good grasping since it enables a mutual synchronisation between the participants. To this end, although there is no force sensing mechanism as such in the hand designed by Pedemonte et al. \cite{pedemonte2016design} and Arns et al. \cite{arns2017design}, they still passively control the closure of the hand based on the deformation, thereby producing a similar synchronous behaviour. Vigni et al. \cite{vigni2019role} observed that the grip strength had an effect on the perception of the robot's personality, which is consistent with the findings of Orefice et al. \cite{orefice2016let}, from human-human handshakes. This can help in crafting behaviours to explicitly yield a personality to the robot, rather than observing such a personality passively. Additionally, encoding different types of such explicit behaviours can help the robot switch to adapt to the human interaction partner if necessary.

\section{Shaking Motions and Synchronisation between Partners}
\label{sec:shaking}
In terms of shaking, one can easily say that there is a synchronisation which takes place between the participants while shaking. This is observed by the studies mentioned above in Sec. \ref{sec:modelling} as well and is one of the aims of most works that study the shaking aspect. They also look at reducing the interaction forces between the robot end-effector and the human hand, which is modelled using impedance/admittance control by some works. 

We divide the works into 3 main categories: Central Pattern Generator (CPG) and Related Models, Harmonic Oscillator Systems and Miscellaneous Shaking Systems, as shown below in Table \ref{tab:shaking}. 

\begin{table*}[]
\centering
\begin{tabularx}{\textwidth}{|>{\centering\arraybackslash}X|c|c|}
\hline
\textbf{Central Pattern Generators and Related Models} & \textbf{Harmonic Oscillator Systems} & \textbf{Miscellaneous Shaking Systems} \\ \hline
 & Beaudoin et al.\cite{beaudoin2019haptic} & \\
Jouaiti et al.\cite{jouaiti2018hebbian} & Chua et al.\cite{chua2010human} & Avraham et al. \cite{avraham2012toward} \\
Kasuga and Hashimoto\cite{kasuga2005human} & Dai et al. \cite{dai2019research} & Karniel et al.\cite{karniel2010turing} \\
Melnyk et al.\cite{artem2013physical}, Melnyk and Henaff\cite{melnyk2016bio} & Mura et al.\cite{mura2020role} &  Nisky et al.\cite{nisky2012three}\\
Papageorgio and Doulgeri \cite{papageorgiou2015kinematic} & Wang et al.\cite{wang2008modelling,wang2009hmm} & Pedemonte et al.\cite{pedemonte2017haptic} \\
Sato et al. \cite{sato2007synchronization}  & Yamato et al. \cite{yamato2008development} & \\
 & Zeng et al.\cite{zeng2012human} & \\ \hline
\end{tabularx}
\caption{Division of works studying the shaking motion. }
\label{tab:shaking}
\end{table*}

\subsection{Central Pattern Generators (CPGs) and Related Models}
Central Pattern Generators (CPGs) \cite{hooper2001central} are biologically inspired neuronal circuits that generate rhythmic output signals. One of the first works to develop an algorithm for the shaking phase proposed the idea of using a CPG-like neural oscillator to model the motion of the shoulder and elbow joints of a robot. They use the torque exerted on the joints as input and generate an oscillatory trajectory, that can be tuned by adjusting gains to amplify the input signal to go from active (high gain) to passive (low gain) \cite{kasuga2005human}.

One drawback of this method, as pointed out by Sato et al. \cite{sato2007synchronization} is that there are quite a few hand-tuned parameters. Therefore, they propose a polynomial approximation for the attractor model of the CPG and subsequently, a model for updating these parameters in an online fashion. A similar on-the-fly parameter update of the oscillator is done by Papageorgiou and Doulgeri \cite{papageorgiou2015kinematic}, who use an impedance model to help tune the parameters of an internal motion generator modelled as a Hopf Oscillator \cite{hopf1942abzweigung}. Although this is not a CPG model, it shows similar synchronisation properties to produce rhythmic outputs like a CPG. The output of the impedance model and oscillators are used to update the oscillator parameters using Direct Least Squares in each iteration using $n$ previous samples of the trajectory. This is unlike previous approaches where the adaptability of the CPG was an inherent trait. 

In contrast, some works directly learn the CPG frequencies to enable a more online real-time approach. Melnyk et al. \cite{artem2013physical} build the CPG around an online learning mechanism that helps it sync with the human's motions directly. Their method dynamically adjusts to changes in the human's shaking frequency and synchronise with the human. Like the above-mentioned works, they too use the joint forces as an input to generate the motions. However, they only work on controlling a single degree of freedom. A similar model is proposed again by Melnyk and Henaff \cite{melnyk2016bio} using two different modes, joint positions and accelerations respectively as the inputs. Like their previous work, they control only one degree of freedom. Along similar lines, Jouaiti et al. \cite{jouaiti2018hebbian} use a similar CPG model and incorporate dynamic plasticity \cite{righetti2006dynamic} in it, making it easier to synchronise with the handshaking frequency.  Moreover, they also propose learning the amplitude of the oscillations along with the frequency, thereby being more adaptive than previous approaches.

\textit{\textbf{Major Findings}}. Overall, CPGs and oscillatory mechanisms synchronise well with the human's motion especially those that are dynamically learnt. They can also fare better than a conventional impedance control approach in terms of flexibility, naturalness, affinity and kindness of the perceived handshake \cite{kasuga2005human}. Additionally, Jouaiti et al. \cite{jouaiti2018hebbian} observed that incorporating plasticity in the CPG can help decrease the energy spent by the robot as well. One major drawback is the ability of such oscillatory mechanisms to converge quickly to the required frequency, taking more than a few seconds even in the fastest cases. This would lead to unnatural handshaking behaviours that take too long to synchronise. Further research is required to increase the convergence speed of such mechanisms. 

\subsection{Harmonic Oscillator Systems}
Harmonic oscillator models are those that employ harmonic systems, like spring-damper systems \cite{dai2019variable,mura2020role,yamato2008development} or simpler sinusoidal motions \cite{beaudoin2019haptic,wang2009hmm,wang2008modelling,zeng2012human} to model the motion during shaking. Some works use both types of harmonic oscillator models in a two-step predictive and reactive system \cite{chua2010human,zeng2012human}. 
Most works that employ harmonic oscillator models use them as reference motions for an impedance controller used to control the joint motions. 

Beaudoin et al. \cite{beaudoin2019haptic} incorporate an impedance controller with different stiffness values using a sinusoidal reference trajectory with different frequencies and amplitudes along with the grasping model proposed by Arns et al. \cite{arns2017design}. Dai et al. \cite{dai2019variable} develop a controller for a custom-made hand that controls the stiffness, viscosity and joint angles independently. Mura et al. \cite{mura2020role} explore different shaking strategies w.r.t. robotic arm stiffness and their synchronisation with the human during handshaking. The parameters of the oscillations are estimated quickly in an online fashion using an Extended Kalman Filter from a fixed number of preceding frames. They compare three models of varying stiffness, namely high, low and variable based on the pressure exerted by the human, similar to Vigni et al. \cite{vigni2019role}. 

Wang et al. \cite{wang2008modelling} propose an impedance control mechanism to model the handshaking mechanism and show how this can be learnt from human handshakes using least-squares minimisation. The reference trajectory for the model is generated using an amplitude of 10cm and mixed frequency components from 0 - 25Hz. Based on the human's response, the model parameters are fine-tuned when the human is being passive. When the human is being active, the model is used to estimate the interaction forces between the human and the robot and carry out the handshake while being passive. This low-level controller is expanded on in \cite{wang2009hmm} where a high-level controller is used to generate reference trajectories for it. They first propose a new method using recursive least squares for a fast online estimation of the impedance parameters which are fed into an HMM that predicts the intention of the human i.e. active or passive from haptic data. The impedance parameters are used, rather than raw force inputs, since they convey the state of the system.

\textbf{\textit{Major Findings}}. The use of active impedance control was found to be more compliant to human motions as compared to simple position-based control \cite{wang2008modelling}. Such active behaviours were rated better in terms of responsiveness than passive ones which also had a significant effect on the perceived synchronisation \cite{mura2020role}. Additionally, it was observed that the human partners would adapt their handshake to the robot's behaviour, even when a change was not explicitly mentioned \cite{dai2019variable}. This further shows the inherent synchronisation that takes place during handshaking, and that we as humans infer it from the interaction itself. There are still no studies that compare the perception of CPG-based shaking motions with harmonic oscillator motions. For a fair comparison, user studies with the same interface would be needed to analyse the perceptual differences between these methods. 

\subsection{Miscellaneous Shaking Systems}
\label{ssec:misc_shaking}
Karniel et al. \cite{karniel2010turing} describe an experimental framework for a Turing test of motor intelligence for shaking behaviours. They do so on a 1D force-controlled haptic stylus, that is presented to a participant. In their test, the forces driving the participant's stylus is a linear combination of forces exerted by an experimenter and different proposed models. They develop a Model Human-Likeness Grade (MHLG) which measure how human-like the motions are from the participants' feedback. Nisky et al. \cite{nisky2012three} extend this to three different versions of the test. The first is a computer vs human test, where the participant is presented wither with a purely algorithmic handshake or a purely human handshake. However, this was not sensitive enough as the participants could almost always guess correctly when a human was shaking their hand. The second version is where the participants have to compare an algorithmic handshake with a noisy human one. The third is the weighted linear combination test proposed by Karniel et al. \cite{karniel2010turing}. Unlike the "pure" test, the authors claim that the latter two variants are said to be better suited for this purpose. 

Avraham et al. \cite{avraham2012toward} make use of this "noise"-based Turing Test to compare 3 different shaking behaviours. The first being a tit-for-tat model that initially records the human's motion passively and then keeps replaying the same motion, assuming that the human's motion stays the same again. The second is a biologically inspired model that simulates a movement that could be generated by extensor and flexor muscles to ensure a low amount of overall interaction force. The final is a simple machine learning model that uses linear regression to learn the parameters of a linear combination of state variables with corresponding Gaussian kernels. It was found that the tit-for-tat model and the machine learning model fare similar to each other. They both fare much better than the biologically inspired model, which the authors argue can be improved by tuning the hyperparameters. While the proposed Handshaking Turing tests work for shaking a simple 1D stylus, it still needs to be seen how well these tests would fare on more complex robotic hardware.

Pedemonte et al. \cite{pedemonte2017haptic} introduce a mechanism for remote handshaking using the hand developed in \cite{pedemonte2016design}. They develop a vertical rail mechanism that the hand is mounted on to support a vertical shaking motion that is passively controlled. The same mechanism is used by both the participants. This shaking motion along with the forces exerted on the hand is relayed to the opponent's hand and rail mechanism to allow a bilateral handshake to take place remotely. They show that their mechanism allows for realistic haptic interaction to take place remotely where the participants can adequately perceive each other's motions and forces.

\section{Human Responses to Social Aspects of Robotic Handshaking}
\label{sec:social}
In previous sections, we have already talked about how some of the different works were perceived in HRI experiments. In this section, we expand further along similar lines and discuss works whose main aim was analysing the responses in such HRI experiments. 

\subsection{External Factors in Handshaking}

While the sense of touch can convey emotional information, there are additional factors that enhance the perception of these feelings and the acceptance of the interaction. Below, we discuss some works that explore different external factors and present their findings to show the importance of fine-tuning these external factors, which although are subtle, have an impact on the way a handshake is perceived. 

Ammi et al. \cite{ammi2015haptic} and  Tsalamlal et al. \cite{tsalamlal2015affective} performed studies to explore how touch influences the perception of facial emotions. They used two haptic behaviours (strong and soft) combined with three visual behaviours namely happy (smiling), neutral and sad (frowning) which were displayed by the robot's lips. They test the interactions in three conditions, haptic-only, visual-only and visuo-haptic. 
The combination of visual expressions with a strong handshake showed higher arousal and dominance over all visual expressions, showing that a sense of touch can enhance robotic expressions. The majority of the comparisons between visuo-haptic and haptic-only cases were insignificant, which the authors argue could be due to the simplistic nature of the facial expression rendering.   Vanello et al. \cite{vanello2010neural} explore similar correlations between participant's perceptions while shaking hands with an artificial hand made of a plastic material while being presented with a visual stimulus of either a human or a robot face. While their experimental design to use fMRI data to understand such correlations is a useful one, their results cannot be deemed as conclusive since only three participants take part in their study.

Nakanishi et al. \cite{nakanishi2014remote} explore social telepresence with a video screen equipped with a robot hand below it. They try out different visibility settings of the participant's hand in the frame and a robotic hand and compare a one-way and a two-way teleoperated handshake settings. They build a hand that resembles a human hand using a soft sponge and gel-like covering and an artificial skin layer to make the appearance more human-like. It is also equipped with resistive wires that heat the fingers. The closure of the fingers is controlled from an external motor with wires connected to the fingertip that are pulled to extend or close the fingers. First, they look at different hand settings of the presenter where their hand would either be visible or out of the video frame (invisible) while shaking. They find that in the invisible case, the interaction was perceived better. Participants not only strongly felt that the presenter was in the same room but also had a strong feeling that they were shaking hands with the presenter in real life. The authors argue that though the visibility of synchronisation might be perceived as better, the visibility of the presenter's hand led to this effect getting cancelled out, which some subjects reported was due to the duplication of the hand i.e. seeing two hands at the same time, both the presenter's and the robot hand. After establishing the results of one-way handshaking, they tested out how participants felt when the presented had a robot hand that was controlled by the participants (two-way handshake). This was tested in two settings where the presenter's interaction was either visible or invisible to the participants. It was seen that the same feelings of physical closeness to the presenter and of shaking hands in real life were rated higher for the invisible two-way case, where participants knew that their handshake was being felt by the presenter. This could possibly be attributed to a perceived synchronisation of sorts which arises from the participant knowing that their actions are being perceived by their partner rather than the interaction being just in a one-way direction.

Jindai et al. \cite{jindai2006development} analyse various handshake motions generated by their model in two ways. They first fit a Bradley Terry Model \cite{bradley1952rank} on paired comparisons of their interactions. Following that, they use a 7 point bipolar scale to test the participant's preferences w.r.t. the handshaking motion, velocity, relief, easiness, politeness and security. They additionally see how participants respond to voice \cite{jindai2007development} and gaze behaviours \cite{jindai2011development}. The study showed that a delay of 0.1 seconds between the voice and handshake motion of the robot was found acceptable. It was found that the most preferred behaviour was when the gaze shifts steadily from the hand while reaching out to the face after contact is established. The response models \cite{ota2014handshake,ota2015handshake} ware tested based on the delay between the request and response motions and it was seen that starting the response a fraction of a second (0.2s to 0.4s) after the response was preferred. However, a larger delay of 0.6s was less preferable. The request model in \cite{jindai2015handshake} was additionally tried out with a human approaching the robot from a distance. They experimented with starting the request at different distances of the human from the robot. Apart from this, both the request and response models were combined with a similar transfer function as in \cite{jindai2006development} such that the robot requests a motion if the human doesn't. This type of behaviour was well perceived by humans as well (positive feedback on a 7-point bipolar scale).

\subsection{Influence on the Perceived Image of the Robot}
As mentioned in the introduction (Sec. \ref{sec:intro}), handshaking can impact first impressions. Therefore in the context of HRI, this can possibly help strengthen the perception of a robot for further interactions that take place. This is explored by the works described below, wherein the effect of robotic handshaking is studied on the specific tasks that a robot has to accomplish.

Avelino et al. \cite{avelino2018power} use their previously proposed handshaking model \cite{avelino2017human} to see how a handshake affects a subsequent interaction wherein the robot needs to perform a navigation task during which, it would need some assistance by the human. They found that participants who shook hands with the robot found it to be warmer and more likeable and were more willing to help the robot for its task. However, they argue that if the robot had an extremely human-like handshake, participants would not anticipate it to get stuck in a simple navigational task due to a mismatch between the behaviour during the experiment and the perceived behaviour according to the handshake. 

Bevan and Fraser \cite{bevan2015shaking} perform an experiment to see the effect of handshaking on negotiations between participants, where one participant interacts with the other via telepresence on a Nao robot. It was seen that handshaking improved mutual cooperation, leading to a more favourable negotiation result for both parties. A haptic feedback for the telepresent negotiator didn't have a significant impact. They also found that handshaking did not affect the degree to which negotiators considered their opponent trustworthy, which they argue is possibly due to the childlike nature of the Nao robot.

\subsection{Distinguishing Ability of Handshakes}
\label{ssec:personality}
In Sec. \ref{sec:modelling}, it was shown that there were observable differences of gender and context on handshaking. In this section, we discuss some works aim to model the certain aspects of the participant, like their gender, personality and mood based on handshakes.

Orefice et al. \cite{orefice2016let} propose a model for making distinctions along the lines of gender and personality (introversion/extroversion), using a set of 20 parameters relating to acceleration, velocity, duration, pressure etc. They find that in male-male pairs, more pressure is applied than in male-female ones. Moreover, they found that female pairs have a longer duration and a lower frequency but the maximum speed of the oscillations is higher. They argue that some results could also be due to the hand sizes rather than gender since most of the females in their study had smaller hands. Coming to personality, they found that introverts reached a higher speed while shaking hands and extroverts would exert more pressure. Though they performed similar experiments in a human-robot handshaking scenario as well, the small number of participants ($n=8$) makes their results in this aspect inconclusive. Garg et al. \cite{garg2017classifying} similarly aimed to classify people's personality into weak and dominant. They use similar information extracted using a custom-made glove to measure accelerations, Euler angles and polar orientations. The features are ranked based on the Mutual Information followed by classification using K Nearest Neighbours, achieving a 75\% accuracy. 
Orefice et al. \cite{orefice2018pressure} perform another longitudinal study with 11 participants over 16 non-consecutive days, that looks at how pressure variations while shaking hands reflect the mood of the participants. They use a custom-made glove with various pressure sensors and an accelerometer worn by both the participant and a Pepper robot. Before shaking hands, participants had to declare which mood (Calm, Relaxed, Cheerful, Excited, Tense, Irritated, Sad, and  Bored) best described their current mood. A consistency in the mood was seen when participants shook hands with a human subject and with pepper, which was unexpected as one would expect an interaction with a robot to seem unrealistic or not as human-like. Overall, no significant results were found for most positive moods, except between "Calm" and "Cheerful" where the former had less pressure observed. In the case of negative moods, "Bored" handshakes had lower pressure than "Excited" and "Tense", which have more arousal than "Bored". In general, lower pressures were found with moods with lower arousal. This shows how handshaking can be used as an affective interaction to further increase the emotional understanding of robots.

\subsection{Human-likeness of Robotic Handshakes}

In the introduction (Sec. \ref{sec:intro}), we mentioned the importance of having human-like body movements, which plays an important role in HRI acceptance \cite{chaminade2005motor,kupferberg2011biological,mori1970uncanny}. Therefore we analyse works that look at the human-likeness of robotic handshakes and draw insights that can help shape future experiments.

The social responses to the shaking models proposed by Wang et al. \cite{wang2008modelling,wang2009hmm} were analysed further in \cite{wang2011handshake,giannopoulos2011comparison}. Both studies perform their experiment on a robot with a rod as its end effector, in a bar like setting wherein participants have noise-cancelling headphones playing bar music and having ambient conversations. In both studies, participants had to perform around 6-7 handshakes in each of the three different handshake settings. First was the basic algorithm proposed in \cite{wang2008modelling}, the second was the interactive model proposed in \cite{wang2009hmm} and third was a human operating the robot. After the handshakes, participants had to rate the human-likeness of the handshake from 1 (resembling a robotic handshake) and 10 (resembling a human handshake). In neither of the studies did the participants see the robot. In the study by Giannopoulos et al. \cite{giannopoulos2011comparison}, participants were first blindfolded and led to the robot whereas, in the study by Wang et al. \cite{wang2011handshake}, participants had a VR headset on which had a graphical rendering of a bar with a human model rendered for the robot, who would walk up and request for a handshake with the virtual hand in the same position as the robot end effector in the real world. In both the studies, the human-operated handshake was rated the highest (6.8/10 in both), followed by the HMM-based handshake (5.9/10 in \cite{giannopoulos2011comparison} and 5.3/10 in \cite{wang2008modelling}). The least was the basic handshake proposed in \cite{wang2008modelling} which was rated much lower than the other 2 alternatives (3.3/10 in \cite{giannopoulos2011comparison} and 3.0/10 in \cite{wang2008modelling}). Although the HMM-based handshake was rated closely as the human-operated one, they both were far from the maximum human-likeness score (10/10), which could be due to the rod-like end effector used. Having a similar experiment with a more sophisticated robot hand could yield the results to be more human-like. 
However, their studies conclude that having an adaptive handshake that matches the behaviour of the human ends up being perceived closer to a human behaviour.

Stock-Homburg et al. \cite{stock2020evaluation} study whether a realistic android robot, that is modelled after a human, with soft silicone skin and pneumatically controlled joints can pass a hardware version of the Turing test.  They have 15 participants blindfolded who have to interact with a human and the robot with their hand stretched out twice in a random order, leading to a total of 4 trials each. Although the robot is built to be as realistic as possible, majority of the humans (11/15) correctly guessed the hand in the first attempt itself and by the last handshake, they were all able to guess the hand correctly. However, they only test a static interaction with the robot. For a better evaluation, comparing a handshake behaviour rather than just a static interaction could show better insights into the modelling of a human-like handshake. 

\section{Discussion}
\label{sec:discussion}
Overall, we have talked about the various works that look into human-robot handshaking, however, due to the differences in hardware and metrics used across different studies, it is difficult to come up with a common benchmark to evaluate these studies. Having said that, some qualitative conclusions can be drawn from analysing these studies. In general, over the different stages of handshaking, an element of synchronisation is present. In the reaching phase, this is seen in terms of the similarity of the requestor's and responder's motions, which is why most works modelling the reaching behaviour draw from human-human interactions. Although such behaviours can be learnt using Reinforcement Learning, human trajectories provide a strong prior to enable the learnt motions to be human-like. Following this, synchronisation in the grasping phase can be observed by matching the strength of the partner and can additionally affect the perceived personality of the robot, which highlights the affective nature of the interaction. Finally, the shaking phase is where the main element of synchronisation can be explicitly measured with the interaction forces between the hands. This depends directly on how well the shaking motion adapts to the that of the partner, leading to low levels of interaction forces as the synchronisation gets better. 
Though synchronisation is a major element of handshaking, in reality, it is difficult to be completely in sync, due to differences in hand shape and size, mental states etc. Therefore, a leader-follower situation can arise in the different stages as well, which could reflect on various personal attributes of the interaction partners.

\begin{figure*}
    \includegraphics[width=\textwidth]{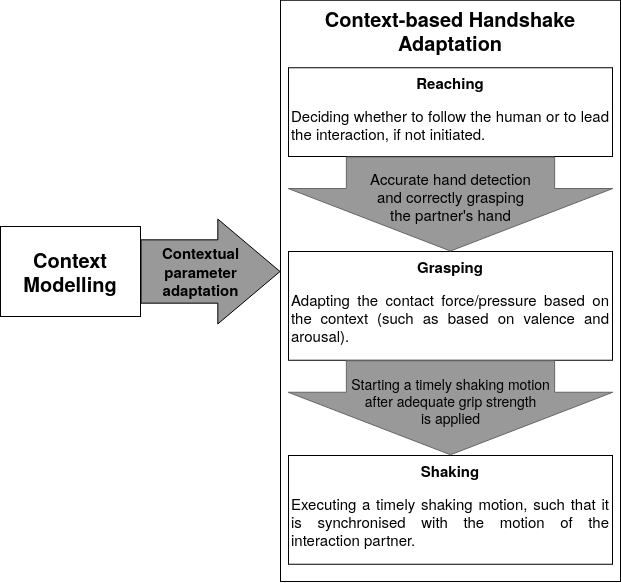}
    \caption{Modified Conceptual Framework for Human-Robot Handshaking. (The proposed suggestions are shaded in grey.)}
    \label{fig:new_framework}
\end{figure*}

Although there is a considerable amount of work on the topic, there are still some gaps in the current state of Human-Robot Handshaking research. Based on what we have already discussed above and their pitfalls, we propose the following suggestions/open areas for further research on Human-Robot handshaking.

\textbf{\textit{Suggestion 1}}: From the perspective of a robotic handshake, making use of contextual cues would be effective in having a successful impact of the handshake. As described in Sec. \ref{sec:modelling} and \ref{ssec:personality}, different social contexts and moods have an effect on the handshake. Being able to detect such cues additionally requires further research in other fields like emotion recognition, intent recognition, etc. Some works try and estimate the mood/personality via the handshake \cite{orefice2018pressure,wang2009hmm}, but there is no explicit context detection in place, say for example by detecting it from facial expressions, or possibly from physiological data, and correspondingly using the insights from Sec. \ref{sec:modelling} for fine-tuning the handshake.

\textbf{\textit{Suggestion 2}}: Developing better social robotic interfaces that have a force sensing mechanism and performing closed-loop control can be more expressive, as seen in \cite{ammi2015haptic,tsalamlal2015affective,vigni2019role}. Currently most works do not implement proper grasping control (as shown in Table \ref{tab:endeff}), which is key for capturing the expressive ability of handshakes to its maximum. Additionally, the human-likeness of an interface is just as important for it's perception. This can be seen in \cite{giannopoulos2011comparison,wang2008modelling} where even though a human was controlling the robot having a rod-like end effector, this mode of shaking only got a human-likeness score of 6.8/10 by the participants.  
Even sophisticated mechanisms, like the Android robot used in \cite{stock2020evaluation} which has a soft skin-like layer and heated palms, are still easily distinguished from a human's hand. One workaround could be to use sensing gloves, like by Orefice et al. \cite{orefice2018pressure}, which could help bridge the gap between a sophisticated interface and social robots.

\textbf{\textit{Suggestion 3}}: One important, yet relatively difficult task is that of combining the different phases. For a more human-like perception, a proper transitioning would be required between each of the different phases. Only a few studies \cite{christen2019guided,jindai2008handshake,mura2020role,pedemonte2017haptic,yamato2008development} look into combining pairs of different phases but still do not implement an end-to-end behaviour. 
There is still work that needs to be done to achieve a complete handshaking behaviour. Along with this, the termination of a handshake is an equally important criterion to make the interaction more socially acceptable. Current works neither take a smooth separation into account nor do they analyse the effects of it. A prolonged handshake or an untimely termination can possibly be perceived as unnatural and can affect the subsequent interaction \cite{nagy2020effects}. Therefore an end-to-end handshaking behaviour should take this into account as well.

\textbf{\textit{Suggestion 4}}: Given that one of the main use-cases of handshakes in shaping first impressions is in business cases, the effect of robotic handshaking in such cases hasn't been properly explored. This is especially important given the use of social robots as front-line employees \cite{jeffrey2016meet}. Bevan and Fraser \cite{bevan2015shaking} study a part in a negotiation context. They look at the impact of robotic handshaking on the impression of the negotiation partner who teleoperated the robot. Moreover, they conduct their study with a Nao robot, which can come across as very childlike and not be taken as seriously in such settings.

While suggestions 2 and 4 are still subjective, suggestions 1 and 3 are areas that can be objectively improved and incorporated to improve existing methods not just for Human-Robot handshaking but would be applicable towards other similar physically interactive behaviours as well. With these two suggestions, a modified framework of Human-Robot handshaking is shown in Fig. \ref{fig:new_framework} (the suggested aspects are shaded in grey). Here the main importance is given to contextual modelling, which influences parameters like strength, speed etc. by adapting them accordingly.


\section{Conclusion}
\label{sec:conc}
Handshaking is a versatile non-verbal interaction which plays important roles in social settings. In this paper, we first draw insights from human-human handshaking regarding timing, trends in the grip strength and the synchronisation of shaking. We then explore the different phases of handshaking, namely reaching, grasping and shaking, while observing a common aspect of synchronisation between the phases. We finally discuss how handshaking affects the way the robot is perceived and propose some directions for future work. However, one thing to keep in mind is that handshaking is just one in so many physical interactions all of which vary over different cultures, age groups, geographic locations, contextual settings etc. 
To this end, learning different physically interactive behaviours, such as hand-claps/high-fives, fist bumps, or a combination of different touch-based interactions, would help improve the perception of the robot. Being able to distinguish and learn such new physically interactive behaviours on the go, building a skill library of sorts, rather than just a single one like handshaking, could improve the sociability of a robot.

\section{Acknowledgements}
The authors would like to thank Dorothea Koert, Niyati Rawal and Xihao Wang for their useful comments that helped make the paper better. 

\section{Compliance, Funding, and Conflict of Interest}

\textit{Compliance with  Ethical  Standards}: The  authors  declare that there are no compliance issues with this research.

\textit{Funding}: This research was funded by the Interdisciplinary Research Forum (Forum Interdisziplin\"are Forschung - FiF) at the Technical University of Darmstadt, the Association of Supporters of Market-Oriented Management, Marketing, and Human Resource Management (F\"orderverein f\"ur Marktorientierte Unternehm-ensf\"uhrung, Marketing und Personalmanagement e.V.), the Leap in Time Foundation (Leap in Time Stiftung), and the Center for responsible digitization (Zentrum Verantwortungsbewusste Digitalisierung -  ZEVEDI).

\textit{Conflict of Interest}: The authors declare that they have no conflict of interest.
%
%
%
\bibliographystyle{spmpsci}
\bibliography{samplepaper}

\end{document}